\documentclass[letterpaper]{article} 
\usepackage{aaai25}  
\usepackage{times}  
\usepackage{helvet}  
\usepackage{courier}  
\usepackage[hyphens]{url}  
\usepackage{graphicx} 
\usepackage{natbib}  
\usepackage{caption} 
\usepackage{algorithm}
\usepackage{algorithmic}
\usepackage{subcaption}
\usepackage{booktabs}
\usepackage{multirow}
\usepackage{amsmath}
\usepackage{amsfonts}

\usepackage{newfloat}
\usepackage{listings}
\DeclareCaptionStyle{ruled}{labelfont=normalfont,labelsep=colon,strut=off} 
\floatstyle{ruled}
\newfloat{listing}{tb}{lst}{}
\floatname{listing}{Listing}
\title{Towards Loss-Resilient Image Coding for Unstable Satellite Networks}
\author{
    Hongwei Sha,
    Muchen Dong,
    Quanyou Luo,
    Ming Lu\thanks{Corresponding Author},
    Hao Chen,
    Zhan Ma
}
\affiliations{
    School of Electronic Science and Engineering, Nanjing University\\  \{hongweisha,dmc,luoquanyou\}@smail.nju.edu.cn, \{minglu, chenhao1210, mazhan\}@nju.edu.cn
}

\usepackage{bibentry}

\begin{document}

\maketitle

\begin{abstract}
{Geostationary Earth Orbit (GEO) satellite communication demonstrates significant advantages in emergency short burst data services. However, unstable satellite networks, particularly those with frequent packet loss, present a severe challenge to accurate image transmission. To address it, we propose a loss-resilient image coding approach that leverages end-to-end optimization in learned image compression (LIC). Our method builds on the channel-wise progressive coding framework, incorporating Spatial-Channel Rearrangement (SCR) on the encoder side and Mask Conditional Aggregation (MCA) on the decoder side to improve reconstruction quality with unpredictable errors. By integrating the Gilbert-Elliot model into the training process, we enhance the model's ability to generalize in real-world network conditions. Extensive evaluations show that our approach outperforms traditional and deep learning-based methods in terms of compression performance and stability under diverse packet loss, offering robust and efficient progressive transmission even in challenging environments. Code is available at \url{https://github.com/NJUVISION/LossResilientLIC}.}
\end{abstract}

\section{Introduction}

{Geostationary Earth Orbit (GEO) satellite communication offers distinct advantages over internet-based or Low Earth Orbit (LEO) satellite (e.g., Starlink) transmission, including wide signal coverage and all-weather synchronization. These benefits make it particularly suitable for navigation, communication in remote areas, and short burst data service, as it is less vulnerable to harsh environments. However, GEO satellite communication often suffers from extremely low available bandwidth, large transmission intervals between packets, and severe \textit{packet loss}~\cite{effective,comparative}, posing significant challenges for emergency image transmission.}

\begin{figure}[t]
\centering
\includegraphics[width=1.0\linewidth]{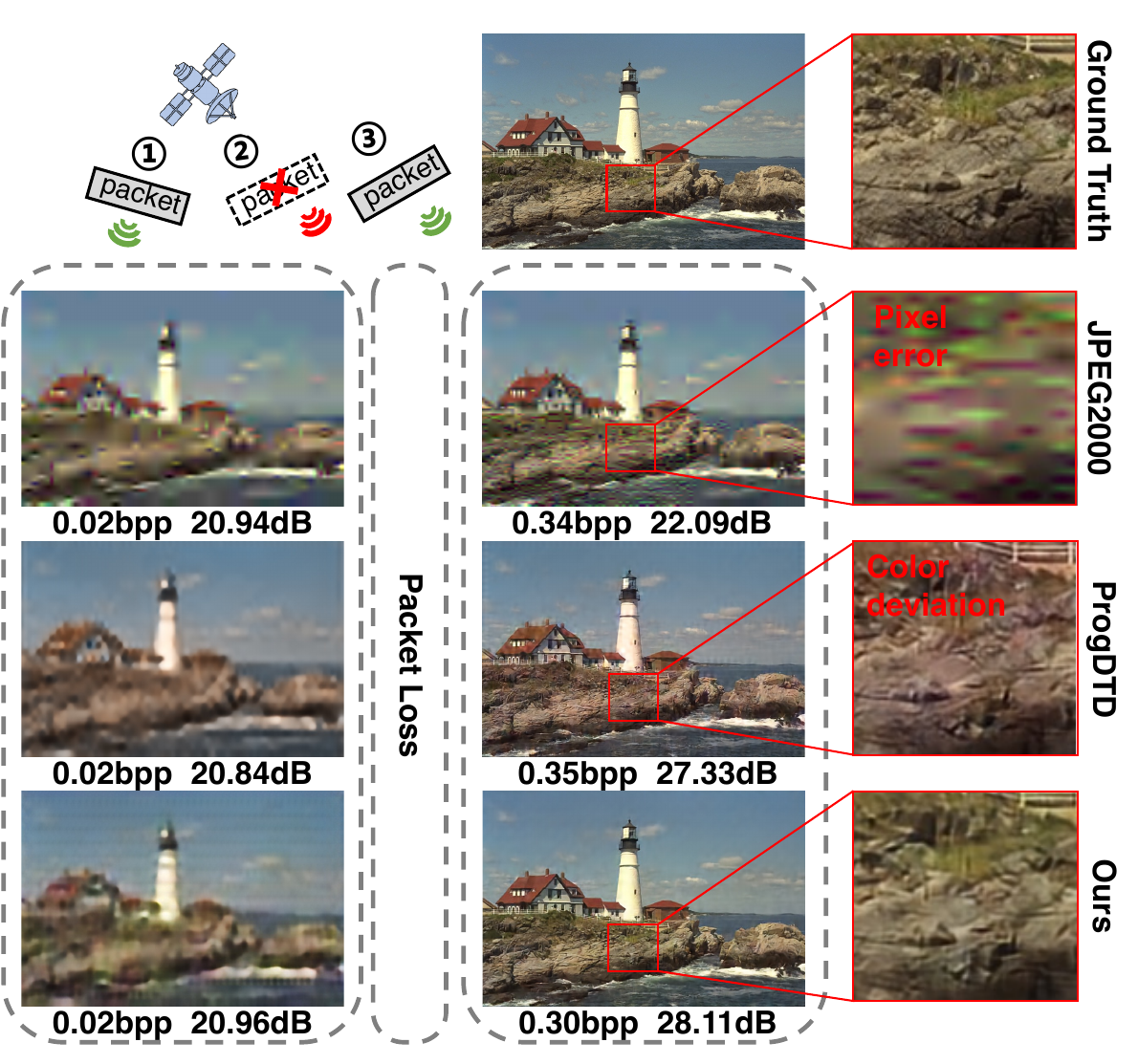}
\caption{{Visualizations of reconstruction results with packet loss. Even under progressive transmission, burst packet loss can disrupt the subsequent reconstruction, leading to unavoidable pixel error in JPEG2000 or color deviation in ProgDTD. Our proposed method demonstrates generalized resilience to unpredictable loss.}}
\label{fig:example}
\end{figure}

{Over the past few decades, image compression technologies have focused on reducing bitrate consumption under bandwidth constraints. From traditional rules-based image coding standards like JPEG~\cite{jpeg}, BPG~\cite{BPG}, and WebP~\cite{webp} to recent learned image compression (LIC) methods~\cite{balle2018,minnen2018,lu2022transformer,liu2023learned}, substantial progress has been made in enhancing compression performance. In light of high network latency, progressive coding (also known as scalable coding) techniques~\cite{jpeg2000,lee2022dpict,progdtd} have been developed to allow image previews using received packets. These algorithms encode images into a single scalable bitstream, improving the quality of the reconstructed image as more complete packets are received. However, currently, few efforts have been made to consider the impact of packet loss on reconstruction results, especially in learning-based methods.}

{Given an encoder-decoder pair typically deployed in transmission systems, the encoder attempts to eliminate redundancy from the input image, creating a compact representation that encapsulates essential information (e.g., brightness or color) for reconstruction at the receiver. This information is distributed across various packets without any intended redundancy or protection mechanism. When packet loss occurs, the stored information is permanently lost, making restoration during decoding impossible. The decoder then faces difficulties in reconstruction when it receives incomplete data, as the missing packets induce a shift in feature distribution that the decoder cannot accommodate. Such unpredictable packet loss can not only lead to a decline in reconstruction quality but also cause substantial decoding errors, such as pixel error or color deviation, which are illustrated in Fig.~\ref{fig:example}.}

{To address these issues, we propose a loss-resilient approach leveraging the end-to-end optimization of LIC. Our framework builds upon the training-aware progressive coding method ProgDTD~\cite{progdtd}, achieving bitrate scalability over the channel dimension in extremely low-bandwidth satellite networks. Given that each channel of encoded features contains distinctive information critical for decoding~\cite{duan2022opening}, we introduce a spatial-channel rearrangement mechanism on the encoder side, which reallocates each channel's information across multiple ones to prevent irreversible errors. At the decoder end, we employ the Mask Conditional Aggregation (MCA) module by extracting the loss mask from the received incomplete features and using it as additional condition information for the decoder along with the received features. This approach enhances reconstruction results through effective loss modeling. Furthermore, we incorporate packet loss simulation using the Gilbert-Elliot model~\cite{gilbert,elliott} into the training process to improve the generalization of our model in real network environments.}

{Extensive evaluations using commonly used datasets demonstrate that our method surpasses both traditional and deep learning-based approaches in compression performance and loss resilience stability under various random packet loss rates. Our algorithm generalizes well in both simulation and real networking evaluations, preventing pixel error or color deviation during progressive transmission. Moreover, our approach introduces an acceptable level of computational complexity, making it practical for real-world applications.}

{Our contributions can be summarized as follows:
\begin{itemize}
    \item We achieve loss-resilient image coding through spatial-channel rearrangement on the encoder side and mask conditional aggregation on the decoder side.
    \item We introduce the Gilbert-Elliot model to guide the training of our model, enhancing its generalization in both simulated and real-world satellite networks.
    \item Experiments demonstrate that our method performs better in various simulated and real-world environments, providing efficient and stable progressive transmissions.
\end{itemize}}

\section{Related Work}
\subsection{Learned Image Coding}\label{sec:LIC}
{Recent advancements in LIC have notably improved the compression performance. Unlike traditional techniques such as JPEG and JPEG2000~\cite{jpeg2000}, neural network-based methods utilize end-to-end training to adaptively derive optimal feature representations, achieving a superior balance between compression ratio and image quality.
Ball\'e et al.~\cite{balle2016} introduced the first end-to-end image compression method leveraging the Variational Auto-Encoder (VAE)~\cite{vae}, which was then enhanced by learning the probability distribution of the latent representation using a hyperprior model~\cite{balle2018}. Furthermore, Minnen et al.~\cite{minnen2018} proposed to establish spatial feature distribution dependencies using the autoregressive model. Although this method further enhanced compression performance, it introduced dependencies in distribution estimation, which complicated the decoding of subsequent features under packet loss. The succeeding works primarily focused on advancing the aforementioned framework, either by employing more sophisticated transform networks~\cite{lu2022transformer,liu2023learned} or designing more efficient context models~\cite{minnen2020channel,he2021checkerboard}, to enhance the compression performance and computational efficiency of LIC.}

{Among them, progressive coding techniques have been developed to improve the user experience, particularly in high-latency network environments. This method allows images to be basically decoded in a coarse reconstruction, which gradually becomes clearer as more bitstream is received. \citeauthor{toderici2017}~\cite{toderici2015,toderici2017} proposed a progressive coding method by utilizing Recurrent Neural Networks (RNNs). \citeauthor{cai2019novel}~\cite{cai2019novel} introduced a method that utilized two scales of latent representations to achieve progressive coding. Hojjat et al.~\cite{progdtd} demonstrated the varying significance of data stored in the latent channels. With the double-tail-drop strategy, prioritized channels were transmitted first, facilitating progressive coding via receiving ordered channels.}

\subsection{Loss-Resilient Methods}
\label{sec:Loss-Resilient Method}

{Loss-resilient methods are designed to address random packet loss during transmission. These methods develop strategies for potential error control or concealment to improve reconstruction quality across the compression and transmission process. The representative methods include Forward Error Correction (FEC)~\cite{LDPC, rateless}, which involves introducing redundancy into the bitstream as a safeguard. While this approach may decrease the coding efficiency, it helps to mitigate the occurrence of errors to some extent. Another solution involves coding adjacent macroblocks independently and dividing them into separate packets, as suggested in~\cite{chu1998detection,ismaeil2000efficient}. This method reduces the impact of packet loss but limits the encoder's ability to exploit redundancy between adjacent macroblocks, resulting in a larger bitstream size. Other methods, such as post-processing at the decoder side, can also help to alleviate packet loss errors. Wang et al.~\cite{zhu1993coding, zhu2} introduced an image restoration algorithm designed to recover the lost packet data, functioning as a low-pass filtering while overlooking detailed information. This approach often yields limited benefits and also adds computational complexity to the decoder.}

{Deep learning-based methods rarely take into account the impact of packet loss on compression model performance. \citeauthor{grace}~\cite{grace} introduced random packet loss during training to help the video codec adapt to feature distribution shifts caused by packet loss. However, this approach only attempted to make the model fit additional data beyond the training set and did not explicitly incorporate loss-resilient structures into the model design. Our proposed method aims to enhance loss resilience in LIC by integrating specific anti-loss strategies. In contrast,~\citeauthor{liu2023bitstream}~\cite{liu2023bitstream} address JPEG corruption through post-repair strategies but significantly increase decoding complexity. Ours enhances loss resilience in LIC by integrating specific anti-loss strategies directly into the codec design, ensuring robust performance in unstable networks with minimal computational overhead}

\section{Proposed Method}
\subsection{Preliminary}
\begin{figure}[t]
\centering
\includegraphics[width=1.0\linewidth]{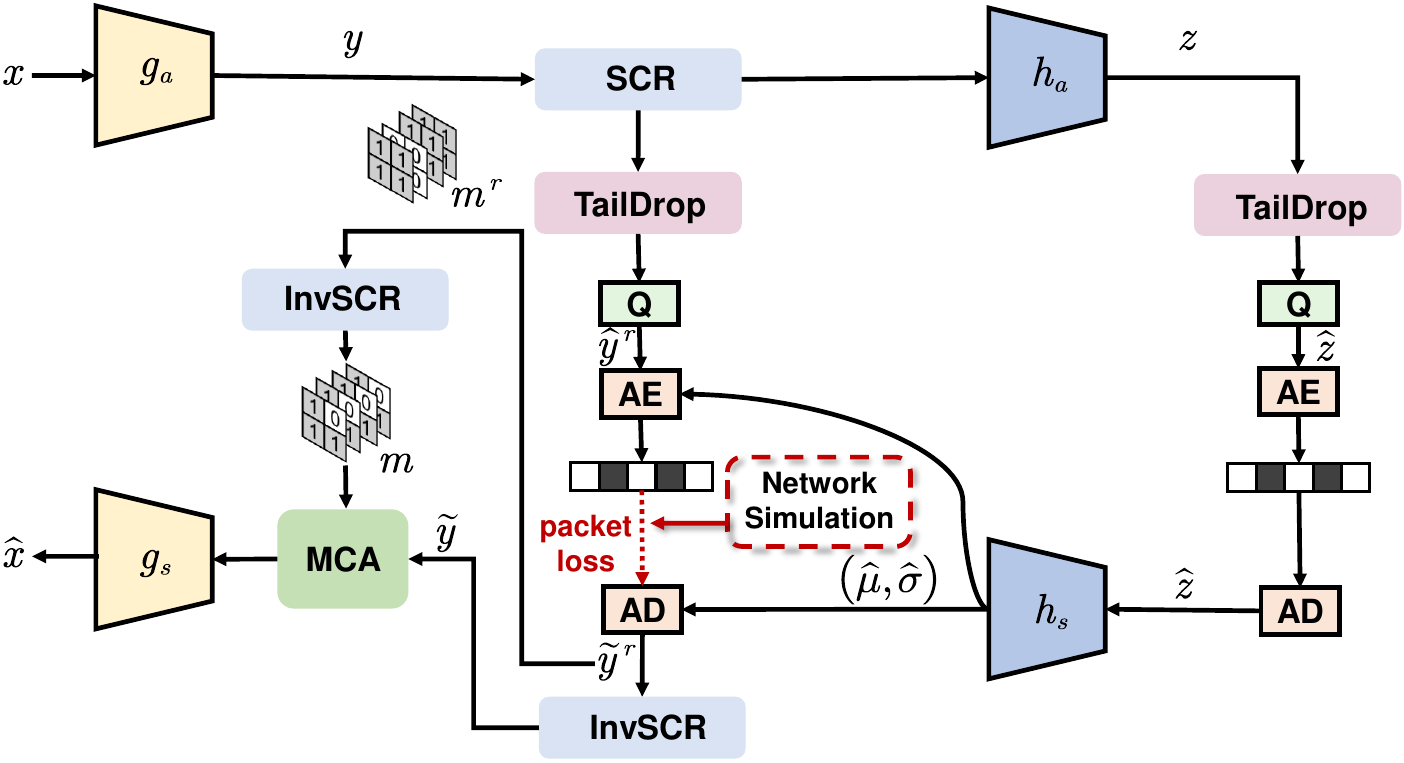}
\caption{{Overview of the architecture of our proposed loss-resilient image coding method. AE and AD represent arithmetic encoding and decoding, respectively.}}
\label{pipeline}
\end{figure}

{We illustrate our proposed loss-resilient method using the milestone LIC framework proposed by \citeauthor{balle2018}~\cite{balle2018}. The overall architecture is depicted in Fig.~\ref{pipeline}.}

{Given an input image ${\bf x} \in \mathbb{R}^{H \times W \times 3}$, the analysis encoder $g_a(\cdot)$ is firstly applied to derive the compact latent representation $\bf y$, which will be quantized (denoted by ``Q'') into $\bf \hat{y}$ for further entropy coding. To obtain an accurate distribution estimation for $\bf \hat{y}$, the hyper encoder $h_a(\cdot)$ and decoder $h_s(\cdot)$ pair is employed to generate the parameters mean $\hat{\mu}$ and scale $\hat{\sigma}$, assuming a Gaussian distribution. The hyperprior variable $\bf z$ is subsequently quantized into discrete representation $\bf \hat{z}$ for entropy coding using the simple factorized model, which is omitted in Fig.~\ref{pipeline} for simplicity. Once the encoded bitstreams of $\bf \hat{y}$ and $\bf \hat{z}$ are properly transmitted, the synthesis decoder $g_s(\cdot)$ can reconstruct the desired output $\bf \hat{x}$. The entire model utilizes a rate-distortion (R-D) loss function for end-to-end optimization by
\begin{align}
\mathcal{L}&=\mathcal{R}({\bf \hat{y}})+\mathcal{R}({\bf \hat{z}})+\lambda \cdot \mathcal{D}({\bf x}, {\bf \hat{x}}) \nonumber \\
&=\mathcal{R}({\bf \hat{y}})+\mathcal{R}({\bf \hat{z}})+\lambda \cdot \mathcal{D}({\bf x}, g_s({\bf \hat{y}}))\text{,} \label{rd_1} \\
&\text{where}~{\bf \hat{y}} = \text{Q}(g_a({\bf x})) \text{.} \nonumber
\end{align}
Here, $\mathcal{R}({\bf \hat{y}})$ and $\mathcal{R}({\bf \hat{z}})$ represent the bitrates consumed by ${\bf \hat{y}}$ and ${\bf \hat{z}}$, respectively, and $\mathcal{D}({\bf x}, {\bf \hat{x}})$ denotes the distortion between the reconstructed image ${\bf \hat{x}}$ and the input image ${\bf x}$. The parameter $\lambda$ acts as a Lagrange multiplier, balancing the trade-off between rate and distortion.}

{To enable progressive coding under the extremely low-bandwidth satellite network, we follow the method in ProgDTD~\cite{progdtd} and adopt the double-tail drop strategy~\cite{taildrop}. By randomly zeroing out the channels at the tail of latent features during training, the model learns to concentrate important information in the earlier channels. This allows for progressive decoding by sequentially transmitting the packaged channels. We adopt this approach because it is simple and efficient, requiring no additional network modules while still achieving remarkable progressive coding performance.}

{Considering potential packet loss issues, we enhance the resilience of the LIC model from both the encoder and decoder perspectives, based on the channel-wise progressive coding pipeline. On the encoder side, we apply Spatial-Channel Rearrangement (SCR) to distribute information from each channel across multiple different channels, thereby preventing the complete loss of crucial information (detailed in Section~\ref{sec:SCR}). On the decoder side, a Mask Conditional Aggregation (MCA) module is employed to model the packet loss condition, allowing the decoder to adapt to the distribution shift caused by incomplete transmitted features (explained in Section~\ref{sec:Mask-Enhanced Feature Fusion}). Moreover, the Gilbert-Elliot model is introduced to simulate the potential loss conditions and enhance model generalization in real network environments (described in Section~\ref{sec:Simulating Packet Loss}).}

\begin{figure}[t]
\centering
\includegraphics[width=1.0\columnwidth]{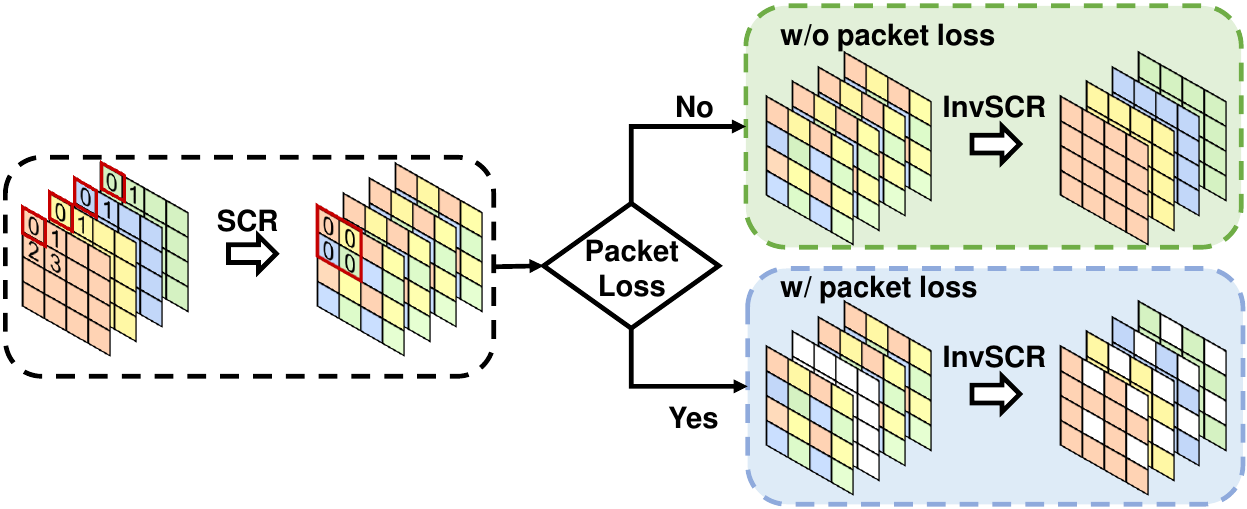}
\caption{{A spatial-channel rearrangement example. Feature points denoted by ``0'' from four channels are rearranged into a grid in a new channel. When the second rearranged channel is lost, the error is distributed across all four channels.}}
\label{fig:SCFR}
\end{figure}

\subsection{Spatial-Channel Rearrangement}
\label{sec:SCR}

{As mentioned earlier, the encoder network $g_a(\cdot)$ is dedicated to eliminating the redundancy from the input image, having a compact representation across the spatial-channel dimension. According to the observations in \cite{duan2022opening}, the diverse channels of the latent features independently contain information such as brightness, color, or edges, which may be unrelated to one another. In the context of channel-wise transmission, as used in this work, if packet loss occurs in a channel that is not correlated with others, it can cause irreversible degradation in the reconstruction result. Therefore, from the encoder perspective, it is essential that each channel has correlated counterparts for error control.}

{To achieve this, we introduce Spatial-Channel Rearrangement (SCR) on the latent feature $\bf y$ to obtain the rearranged feature ${\bf \hat{y}}^r$ losslessly, which is depicted in Fig.~\ref{pipeline}. Taking four adjacent channels in Fig.~\ref{fig:SCFR} as an example, we arrange the feature points at the same spatial location into a $2 \times 2$ grid. This grid is first filled into the top-left corner of the rearranged channel, and then continues sequentially until the four new channels are fully populated. This process distributes the features from each original channel across four new channels, thereby enhancing inter-channel correlation. When packet loss does not occur, applying the inverse SCR (InvSCR) on the decoder side does not change the input feature to the decoder. If any one channel is lost, its information can still be reconstructed using the remaining three channels, effectively minimizing unrecoverable errors.}

{Following the LIC procedure, hyper features are then extracted using the rearranged latent ${\bf \hat{y}}^r$ for distribution modeling. Although our proposed SCR modifies the original spatial relationships of the latent features, it introduces a degree of spatial decorrelation. This facilitates the hyperprior under the assumption of independent and identical distribution and does not adversely affect overall coding efficiency~\cite{ali2024towards}, which can be demonstrated in Section~\ref{ablation}.}

{Due to complexity considerations, this work selects four sequential channels as the process window for SCR. As an extreme case, during the practical packetization process, we will check the last channel in each packet. If the channel number is a multiple of 4, we move it to the next packet. This approach reduces the impact of packet loss without compromising the performance of progressive coding.}

\begin{figure}[t]
\centering
\includegraphics[width=0.7\columnwidth]{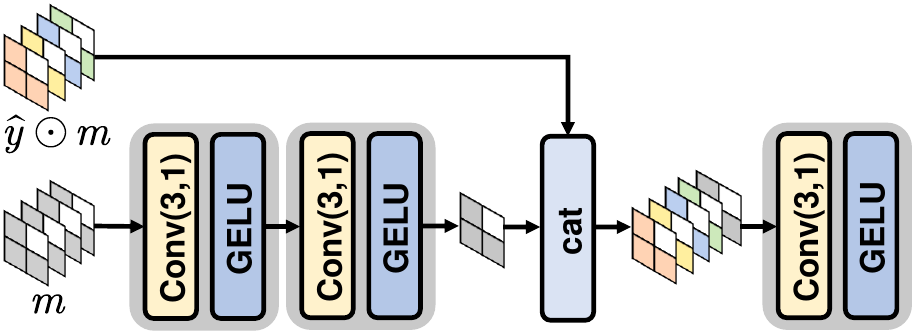}
\caption{{Pipeline of mask conditional aggregation module.}}
\label{fig:MCA}
\end{figure}

\subsection{Mask Conditional Aggregation}
\label{sec:Mask-Enhanced Feature Fusion}

{The key to the success of deep learning-based image compression algorithms lies in their ability to learn general encoding and decoding transformations through end-to-end optimization, based on the distribution of the training data. However, unknown packet loss can cause a shift in the distribution of the latent features received by the decoder, leading to an inability to accurately reconstruct the intended result. Even with the addition of random mask training as in \cite{grace}, the model still cannot learn to adapt to the dedicated packet loss, especially in complex real-world environments.}

{To this end, we propose a Mask Conditional Aggregation (MCA) module on the decoder side to enhance the ability to adapt to the feature distribution with packet loss. Suppose the rearranged latent feature ${\bf \hat{y}}^r$, we employ a binary mask ${\bf m}^r \in \{0,1\}^C$ to denote which channel in ${\bf \hat{y}}^r$ has been lost during transmission. ${\bf m}^r_i = 1$ indicates ${\bf \hat{y}}^r_i$ is received, and ${\bf m}^r_i = 0$ means it is missing ($1 \leq i \leq C$). As a result, the received incomplete feature can be derived by 
\begin{align}
    \bf \widetilde{y}^r = {\bf \hat{y}^r \odot {\bf m}^r} \text{,}
\end{align}
where $\odot$ refers to element-wise multiplication. After applying InvSCR, we can obtain the original arrangement of the latent feature $\widetilde{\bf y}$ and its corresponding loss mask ${\bf m}$.}

{To aggregate the incomplete feature conditioned on the packet loss, we deploy a simple two-layer convolutional network followed by GELU~\cite{hendrycks2016gaussian} activations to model the mask ${\bf m}$. This mask is then concatenated with $\widetilde{\bf y}$ to obtain the combined feature. An extra convolutional layer with the GELU activation is used for feature fusion to input into the decoder. With the incorporation of MCA, the R-D loss can be extended from Eq.~\eqref{rd_1} by
\begin{align}
\mathcal{L}&=\mathcal{R}({\bf \hat{y}^r})+\mathcal{R}({\bf \hat{z}})+\lambda \cdot \mathcal{D}({\bf x}, g_s(M({\bf \widetilde{y}}, {\bf m}))) \text{,} \\ \nonumber
&\text{where}~{\bf \widetilde{y}}={\bf \hat{y} \odot {\bf m}} \text{.}
\end{align}
$M$ responds to the MCA module. By explicitly modeling the mask, the decoder can aggregate features based on specific packet loss conditions, thereby adapting more effectively to feature distribution shifts caused by packet loss.
}

\begin{figure}[t]
\centering
\includegraphics[width=0.8\columnwidth]{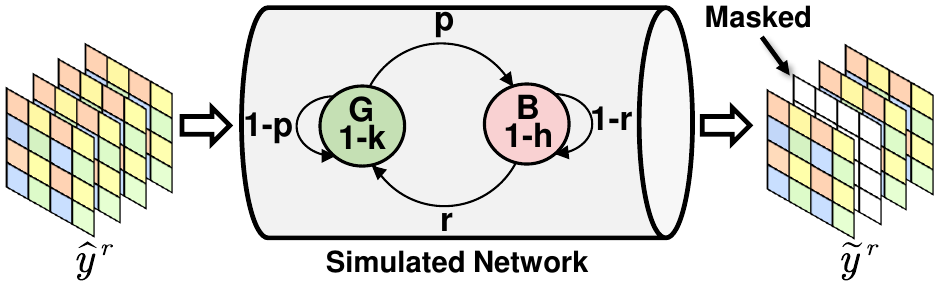}
\caption{{Network simulation using Gilbert-Elliot model.}}
\label{gemodel}
\end{figure}

{It is worth noting that as long as the bitstream for the hyperprior ${\bf \hat{z}}$ is fully received during transmission, the distribution estimate will remain consistent with the encoder side. This ensures the entropy decoding can proceed properly on the decoder side, even if the packets of ${\bf \hat{y}}^r$ are not completely transmitted. As a result, in this work, it is necessary to ensure the lossless transmission of the hyperpriors.}

\begin{table*}[t]
\centering
\begin{tabular}{cc|ccc|ccc|cccc}
\toprule
\multirow{2}{*}{$p_e$} & \multirow{2}{*}{Method} & \multirow{2}{*}{bpp} & \multicolumn{2}{c|}{PSNR(dB)} & \multirow{2}{*}{bpp} & \multicolumn{2}{c|}{PSNR(dB)} & \multirow{2}{*}{bpp} & \multicolumn{2}{c}{PSNR(dB)} \\
\cline{4-5} \cline{7-8} \cline{10-11}
& & & $mean$ \(\uparrow\) & $var$ \(\downarrow\) & & $mean$ \(\uparrow\) & $var$ \(\downarrow\) & & $mean$ \(\uparrow\) & $var$ \(\downarrow\) \\
\midrule
&JPEG2000 & 0.137 & 24.201 & 0.133 & 0.205 & 24.517 & 0.269& 0.342 & 25.013 & 0.310  \\
\multirow{2}{*}{5\%}&Baseline & 0.133 & 26.551 & 0.392 & 0.215 & 26.976 & 0.397 & 0.336 & 28.619 & 0.991 \\
&ProgDTD & 0.133 & 26.312 & 0.436 & 0.221 & 26.877 & 0.820 & 0.359 & 27.919 & 0.564 &\\
&Random & 0.133 & 26.657 & 0.064 & 0.220 & 27.754 & 0.061 & 0.351 & 29.256 & 0.151  \\
&Ours & 0.131 & \textbf{26.850} & \textbf{0.015} & 0.210 & \textbf{28.247} & \textbf{0.043} & 0.338 & \textbf{29.561} & \textbf{0.007} \\
\midrule
&JPEG2000 & 0.137 & 23.448 & 0.220 & 0.205 & 23.626 & 0.226 & 0.342 & 23.966 & 0.318  \\
\multirow{2}{*}{10\%}&Baseline & 0.133 & 25.131 & 0.485 & 0.215 & 25.855 & 0.195 & 0.336 & 27.056 & 0.803 \\
&ProgDTD& 0.133 & 24.701 & 0.979 & 0.221 & 25.695 & 1.007 & 0.359 & 26.711 & 0.604 \\
&Random & 0.133 & 25.858 & 0.195 & 0.221 & 27.320 & 0.156 & 0.351 & 28.569 & 0.041  \\
&Ours & 0.131 & \textbf{26.342} & \textbf{0.061} & 0.210 & \textbf{27.605} & \textbf{0.100} & 0.342 & \textbf{28.801} & \textbf{0.007} \\
\midrule
&JPEG2000 & 0.137 & 22.530 & 0.304 & 0.205 & 22.524& 0.313& 0.342 & 22.564 & 0.337  \\
\multirow{2}{*}{20\%}&Baseline & 0.133 & 23.257  & 1.222  & 0.215 & 22.524 & 1.666 & 0.336 & 23.592 & 1.341\\
&ProgDTD&0.133 & 23.625 & 0.518 & 0.221 & 23.740 & 1.258 & 0.359 & 23.919 & 1.913 \\
&Random &  0.133 & 25.217 & \textbf{0.071} & 0.228 &  25.886 & 0.192 & 0.358 & 27.620 &  \textbf{0.016} \\
&Ours & 0.131 & \textbf{25.578} & 0.108 & 0.217 & \textbf{26.617} & \textbf{0.043} & 0.346 & \textbf{27.966}  & 0.043  \\
\bottomrule
\end{tabular}
\caption{{Performance comparison of models under packet loss with a uniform distribution, where $p_e$ represents the probability of packet loss, $mean$ and $var$ respectively represent the average PSNR and variance for 10 test samples at each bitrate. The bold values indicate the best results. The symbol $\downarrow$ signifies that a smaller value is better for that metric, while $\uparrow$ indicates that a larger value is preferable.}}
\label{table:uniform}
\end{table*}

\subsection{Packet Loss Simulation}
\label{sec:Simulating Packet Loss}

{So far, we have proposed loss-resilient algorithms for both the encoding and decoding sides. End-to-end simulated training has been employed to enhance the model's resistance to packet loss. However, in real transmission environments, packet loss conditions are often complex and unpredictable, deviating from the uniform random distribution assumed in \cite{grace}. Therefore, by more accurately modeling these packet loss conditions, we can guide the model's simulated training to achieve better generalization with real-world scenarios.}

{To address it, we introduce the Gilbert-Elliott (GE) model, a two-state Markov method, to simulate the real-world network environment. This model is widely used to characterize error patterns in networks with lossy connections and to analyze the performance of error detection and correction techniques~\cite{hasslinger2008gilbert}. As shown in Fig.~\ref{gemodel}, the symbol $G$ represents the network being in a ``Good'' state, while the symbol $B$ denotes the network being in a ``Bad'' state. Both states can independently cause errors, with each error occurring as a separate event. We denote the state-dependent error rates as $1-k$ in state $G$ and $1-h$ in state $B$. Having the state at any given time $t$ denoted by $q_t$, the transition probabilities between continuous states are represented by $p$ and $r$:
\begin{align}
    p = P(q_t=B \mid q_{t-1}=G)
    \label{p} \\
    r = P(q_t=G \mid q_{t-1}=B)
    \label{r}
\end{align}
This implies that the network's condition at a previous moment influences its condition at the current, which is intuitively reasonable in real-world environment.}

{In real transmission environments, when performing channel-wise progressive coding, packet loss conditions are difficult to simulate but often exhibit a sustained period. This means that if one channel is lost, the probability of losing the subsequent channel also increases. The GE model easily accommodates such simulations, whereas a random mask cannot, significantly degrades its loss resilience in actual network environments. On the other hand, considering that packetization is not performed during the training process, by which channels are grouped and transmitted based on bitrate constraints during transmission. Consequently, the probability of losing multiple channels within the same group increases significantly. This scenario aligns with the assumption of GE model and better reflects the potential packet loss conditions that may occur.}

{To sum up, the overall training objective is described as:
\begin{align}
\mathcal{L}&=\mathcal{R}({\bf \hat{y}^r})+\mathcal{R}({\bf \hat{z}})+\lambda \cdot \mathcal{D}({\bf x}, g_s(M({\bf \widetilde{y}}, {\bf m}))) \text{,} \label{eq:GELR_cost} \\ \nonumber
&\text{where}~{\bf \widetilde{y}}={\bf \hat{y} \odot {\bf m}} \text{,}~{\bf m} \sim P({\bf m}) \text{.}
\end{align}
$P({\bf m})$ represents the distribution function that models the packet loss, which can either be a uniform distribution or simulated distribution by the GE model.}

{To distinguish between the tail-drop method in progressive coding and the zeroing strategy used to simulate network packet loss, different approaches are taken during the training and inference stages. During training, the tail-drop method randomly selects a certain number of tail channels to zero out in each iteration, while packet loss simulation zeros out any channel based on the distribution model's selection. In the practical inference stage, the tail-drop method directly removes channels that do not need to be transmitted at the current level, whereas packet loss evaluation zeros out the channels corresponding to the lost packets. These two approaches are not in conflict with each other.}

\section{Experimental Results}

\subsection{Implementation Settings}

\subsubsection{Datasets}

{We use the Flicker2W dataset~\cite{liu2020unified} for training which will be randomly cropped into the size of $256\times256$. We perform the evaluation experiments using the Kodak dataset~\cite{Kodak} with the resolution of $768\times512$, which is the size acceptable for extremely low-bandwidth transmission. To thoroughly validate the effectiveness of our method, we also provide the experiments on the dataset with 2K resolution in appendix.}

\subsubsection{Training details}

{We select the ``mbt-mean'' model~\cite{balle2018} implemented in the open-source CompressAI library~\cite{compressai} as our baseline model. We also provide a higher-performance version in appendix. Three lower bitrate points are selected to cover the extremely low bandwidth range of satellite networks, corresponding to lambda values of 0.0018, 0.0035, and 0.0067 when using Mean Squared Error (MSE) optimization. The reason to choose MSE optimized model is that emergency scenarios require a high level of image fidelity.}

{During training, we need to simultaneously simulate tail-drop loss in progressive coding and network packet loss. The tail-drop rate is randomly selected within the range [0, 1.0] as described in \cite{progdtd}. To simulate varying packet loss rates with moderate fluctuations in practice~\cite{grace}, we randomly select values from [0.5\%, 1.5\%, 2.5\%, 3.5\%, 5\%] to approximate the 5\% packet loss rate, from [1\%, 3\%, 5\%, 7\%, 10\%] to simulate the 10\%, and from [2\%, 6\%, 10\%, 14\%, 20\%] to simulate the 20\%. For real-world satellite network simulation during training, we set the GE model parameters [$p$, $r$, $h$, $k$] to [0.378, 0.883, 0.810, 0.938], as derived from \cite{pieper2023relationships}, to simulate a 10\% packet loss rate, based on our practical evaluations.}

{All training is conducted on an NVIDIA RTX 3090 GPU, with 200 epochs and a fixed batch size of 16. We choose Adam as the optimizer and the learning rate is set at $10^{-4}$.}

\subsubsection{Testing Setup}
{The testing involves a specific packetization process. We have predefined a maximum packet size of 900 bytes according to commercial short burst data design. Using our packetization strategy (refer to appendix), channel information within the bitrate limit is placed in a single packet, representing a layer of progressive transmission.}

{As mentioned earlier, the hyperprior bitstream, which constitutes only around 6\% of the entire bitstream, serves as the foundational information and will not experience packet loss in our evaluations, ensuring consistency in the entropy model between encoding and decoding. For fairness, we also ensure that the base layer of JPEG2000 in our comparison method does not experience packet loss, as this would prevent proper reconstruction in subsequent steps.}

\subsection{Evaluation}

\subsubsection{Performance Under Packet Loss}

{We first evaluate loss resilience in a network with uniformly distributed packet loss, comparing performance at 5\%, 10\%, and 20\% loss rates (Table~\ref{table:uniform}). The "Baseline" refers to the "mbt-mean" model, while "Random" uses the baseline trained with random channel masking. Our model achieves significantly better reconstruction quality (higher average PSNR) and greater stability (lower variance) compared to others.}

\begin{figure}[t]
\begin{center}
\begin{subfigure}[b]{0.45\linewidth}
\includegraphics[width=\linewidth]{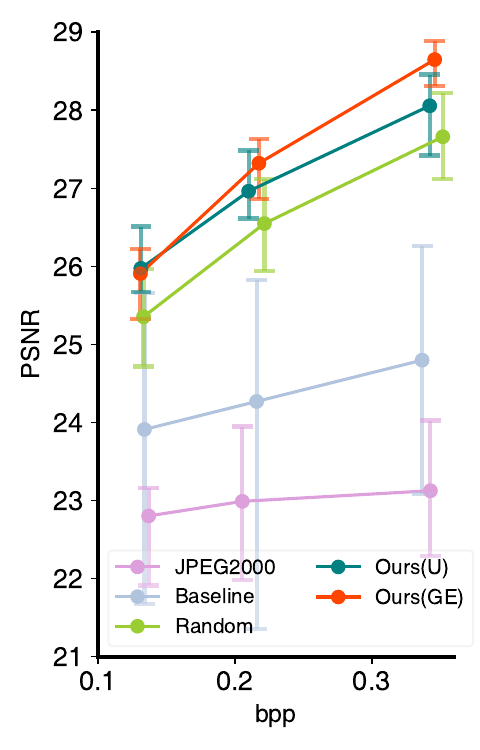}
\caption{GE model with $p_e=15\%$}
\label{fig:ge10}
\end{subfigure}
\hfill
\begin{subfigure}[b]{0.45\linewidth}
\includegraphics[width=\linewidth]{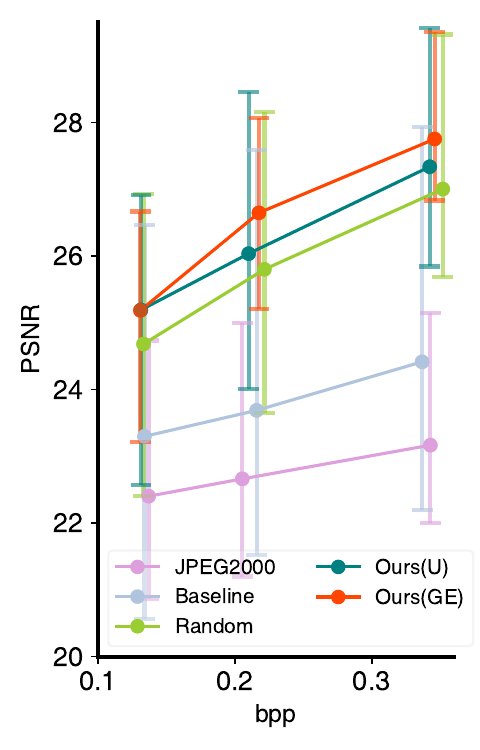}
\caption{GEO satellite network}
\label{fig:geo}
\end{subfigure}
\end{center}
\caption{{Performance under the GE model and the real-world GEO satellite network. The line graph displays the average RD performance over 10 evaluated samples, with the error bar indicating the PSNR range. ``U'' refers to training with the uniform distribution, while ``GE'' indicates training with the GE model to simulate the packet loss.}}
\end{figure}

{Subsequently, we evaluate our model in a simulated network using the GE model. we adjust the packet loss rate to 15\% during testing by setting [$p$, $r$, $h$, $k$] to [0.417, 0.973, 0.620, 0.948], even though our model is trained with a 10\% GE model. This adjustment helps verify the method's generalization after optimization with the GE model. As shown in Fig.\ref{fig:ge10}, our model outperforms others in both performance and stability, especially compared to models trained with uniform packet loss. Additionally, testing on a real GEO satellite network (Fig.\ref{fig:geo}) confirms its superior performance and generalization, highlighting the importance of realistic network modeling for improved loss resilience.}

\begin{figure*}[htbp]
\centering
\includegraphics[width=0.85\linewidth]{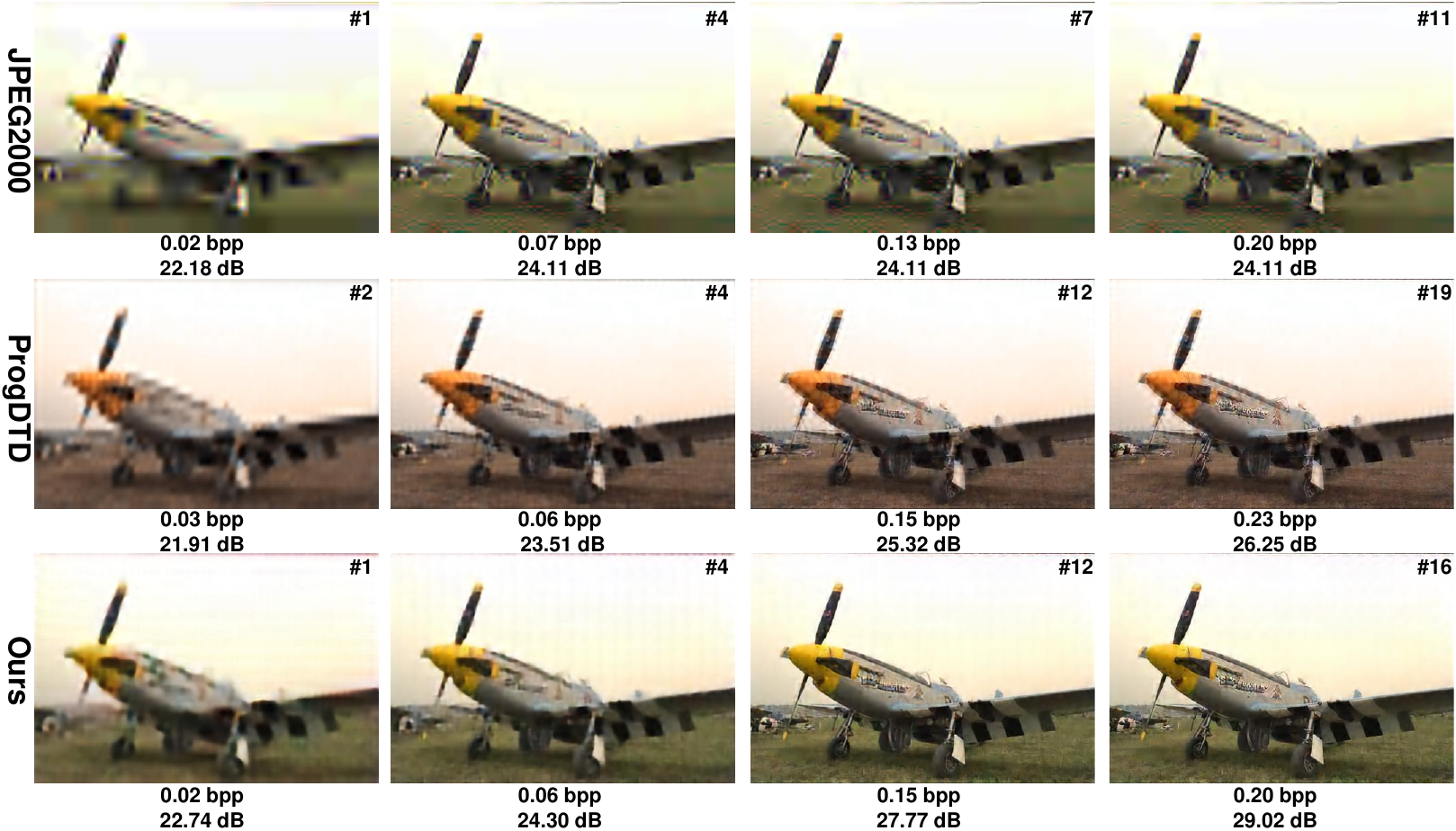}
\caption{{Visualizations of progressive coding under packet loss on Kodim20. \#p denotes the reconstructed result after receiving the p-th packet at the decoder. The first column shows the initial results of progressive transmission without any packet loss for the three methods. The second column shows the results when the 3rd packet is lost. The third column shows the results when the 6th packet is also lost (for JPEG2000, only the 3rd packet is lost). The fourth column shows the decoded results after the progressive transmission is completed with packet loss. Our method consistently demonstrates superior performance.}}
\label{fig:progsub}
\end{figure*}

\subsubsection{Progressive Coding Under Packet Loss}

{Previous experiments only validate the impact of packet loss on reconstruction quality. However, by introducing the progressive coding under bitrate constraint, packet loss can lead to more unpredictable issues. We visualize the reconstruction results of three progressive transmission schemes under random packet loss, i.e., JPEG2000, ProgDTD, and our proposed method in Fig.~\ref{fig:progsub}. For JPEG2000, packet loss prevents correct decoding of subsequent packets, keeping reconstruction quality low despite receiving more packets. Although ProgDTD can still improve reconstruction quality during progressive transmission, the loss of critical color information (packet \#3 in this example) leads to severe color deviation. In contrast, our method demonstrates generalized progressive decoding and significantly outperforms the comparison methods in terms of compression performance. More visualization results can be found in the appendix.}

\subsubsection{Complexity}

{Table~\ref{table:cspeed} reports the parameters, as well as the encoding and decoding time of our method. Due to the additional SCR introduced at the encoding end, encoding time has increased slightly. At the decoding end, incorporating the MCA network has resulted in a modest increase in the total number of parameters, which also leads to a slight extension in decoding time along with the InvSCR. We believe that the minor increase in complexity is acceptable in practice, and as the parameter count of the compression model increases, the additional complexity introduced by our method will become negligible.}
\begin{figure}[htbp]
\begin{center}
\begin{subfigure}[htbp]{0.4\linewidth}
\includegraphics[width=\linewidth]{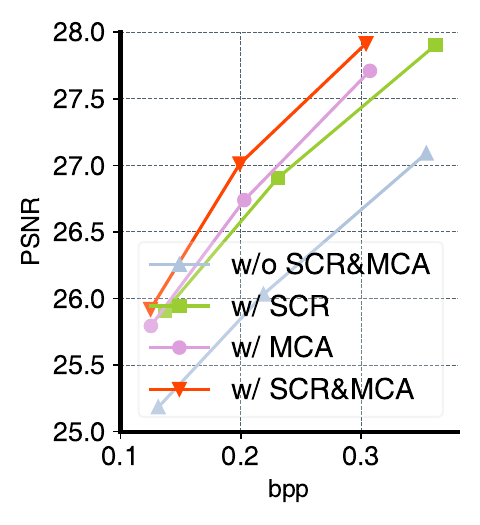}
\caption{Gains of each module.}
\label{fig:module_ablation}
\end{subfigure}
\hfill
\begin{subfigure}[htbp]{0.4\linewidth}
\includegraphics[width=\linewidth]{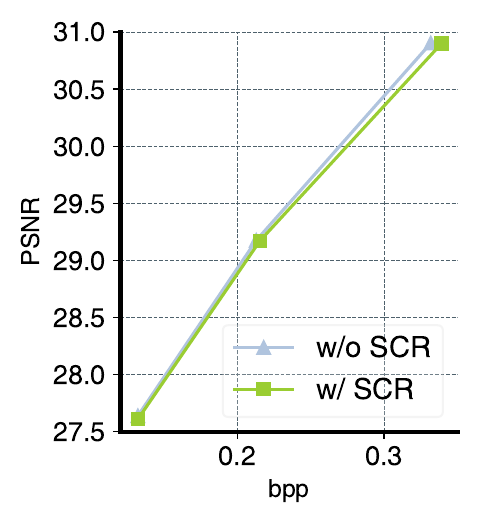}
\caption{The impact of SCR.}
\label{fig:SCR_ablation}
\end{subfigure}
\caption{Ablation studies on modular contributions.}
\end{center}
\end{figure}

\subsection{Ablation Studies} 
\label{ablation}

{We examine the performance gains of each module by integrating them individually into the progressive training, under a 10\% packet loss. As shown in Fig.~\ref{fig:module_ablation}, each module substantially enhances the performance, with the MCA module demonstrating an even greater advantage. Furthermore, the combination of modules further improves the loss resilience of the model.}
{We also verify from Fig.~\ref{fig:SCR_ablation} that SCR has minimal impact on overall coding performance. Although SCR disrupts the original feature positions, it does not substantially affect distribution modeling and instead provides significant improvements in loss resilience.}
\begin{table}[htbp]
\centering
\begin{tabular}{ccc}
\toprule
Method & Params (MB) & EncT / DecT (ms) \\
\midrule
w/o our method & 26.80 & 89.15 / 39.30 \\
w/ our method & 28.53 & 93.35 / 50.17 \\
\bottomrule
\end{tabular}
\caption{{Comparison of model complexity before and after integrating our loss-resilient method. ``Params'' denotes model size. ``EncT'' and ``DecT'' respectively represent the encoding and decoding time to process an image in Kodak.}}
\label{table:cspeed}
\end{table}

\section{Conclusion}

{This paper proposes a loss-resilient LIC method tailored for unstable satellite networks. It enhances compression performance under packet loss through the encoder-side SCR mechanism and decoder-side MCA module. By integrating the GE model to simulate packet loss, our method demonstrates generalization on both simulated and real-world GEO satellite networks. Our work explores a simple and efficient network adaptation method from the perspective of source coding, which has a positive impact on the application of LIC in practical communication scenarios. However, it still cannot address packet loss issues under distributional dependencies, meaning that the loss of any packet affects the subsequent probability estimation. Future work will focus on exploring the potential of LIC methods integrated with autoregressive context models for loss resilience.}

\section*{Acknowledgements}
This work was supported in part by Natural Science Foundation of China (Grant No. 62401251, 62431011 and 62471215), Natural Science Foundation of Jiangsu Province (Grant No. BK20241226), and Jiangsu Provincial Key Research and Development Program (Grant No. BE2022155). The authors would like to express their sincere gratitude to the Interdisciplinary Research Center for Future Intelligent Chips (Chip-X) and Yachen Foundation for their invaluable support.

\bibliography{aaai25}

\clearpage
\appendix

\section{Appendix}
\subsection{Assumptions of Data Packet Transmission}
Our model is developed based on the premise that $\hat z$ will not experience any packet loss during transmission. This is because the encoded hyperpriors $\hat z$ contains all the distribution information (such as mean and scale) of the latent representation $\hat y$, which is critical for correctly decoding. Any loss of $\hat z$ at any position would result in an irreversible decoding error for the corresponding position in $\hat y$. Based on this premise, in practical applications, we need to implement mechanisms such as redundant transmission (e.g., repeating transmission and automatic repeat request) to ensure the accurate and complete delivery of $\hat z$. Fortunately, $\hat z$ constitutes a very small portion of the entire bitstream (typically around 6\% of the overall bitrate for an image), allowing simple and effective error protection with minimal bitrate overhead. Meanwhile, in research on progressive coding, $\hat z$ is also protected as a base layer to ensure that subsequent progressive layers can be decoded correctly, which aligns with the intent of our work.

In summary, as a loss-resilient progressive coding approach, we must ensure that $\hat z$ is transmitted without any loss. The measures to achieve this are both feasible and effective, with minimal cost.

\subsection{Packetization Strategy}
First, we estimate the information content of each channel $\hat{y}_i^r$ ($1 \leq i \leq C$) in $\hat{y}^r$. We then combine the channels in order of their channel indices $i$. When adding the current channel causes the current packet size to exceed the maximum packet size $B_{max}$, we consider the channels before the current channel as one packet.
During actual transmission, there may be slight deviations from the estimated information content. If the actual compressed size of the channels in this packet still exceeds $B_{max}$, we will split the current packet evenly into two packets for transmission.
After completing these steps, only our scheme will also check whether the index of the last channel in the current packet is divisible by 4. If so, we move the last channel of the current packet to the next packet. The pseudocode is shown in Algorithm~\ref{al:packet}.

\subsection{Gilbert-Elliot Model}
The Gilbert-Elliot model is a network simulation system consisting of a Good and a Bad state, with parameters $k$ and $h$ (where $0 \leq k, h \leq 1$) representing the probability of no errors occurring in each state. The condition $h < k$ ensures higher error rates in the Bad state. Transitions between these states occur with probabilities $p$ (Good to Bad) and $r$ (Bad to Good). During each iteration, the GE model outputs a Boolean indicating packet loss: True results in zeroing the corresponding $\hat{y}$ channel data, while False leaves it unchanged. 

This model is particularly suitable for simulating the real situation of satellite networks for two reasons. Firstly, compared to a predefined distribution, it can adjust the various parameters $p,r,h,k$ to emulate various network conditions, allowing for flexible and realistic modeling. Secondly, it effectively simulates continuous burst packet losses. For instance, satellite networks often experience prolonged periods of poor connectivity due to factors like adverse weather, leading to sudden spikes in packet loss. The model’s ability to represent these "Bad" states aligns well with the real-world challenges faced in satellite communications, ensuring our model, trained in such simulated networks, can achieve better compatibility with continuous packet loss in real environments.
\begin{figure}[t]
\centering
\includegraphics[width=0.8\columnwidth]{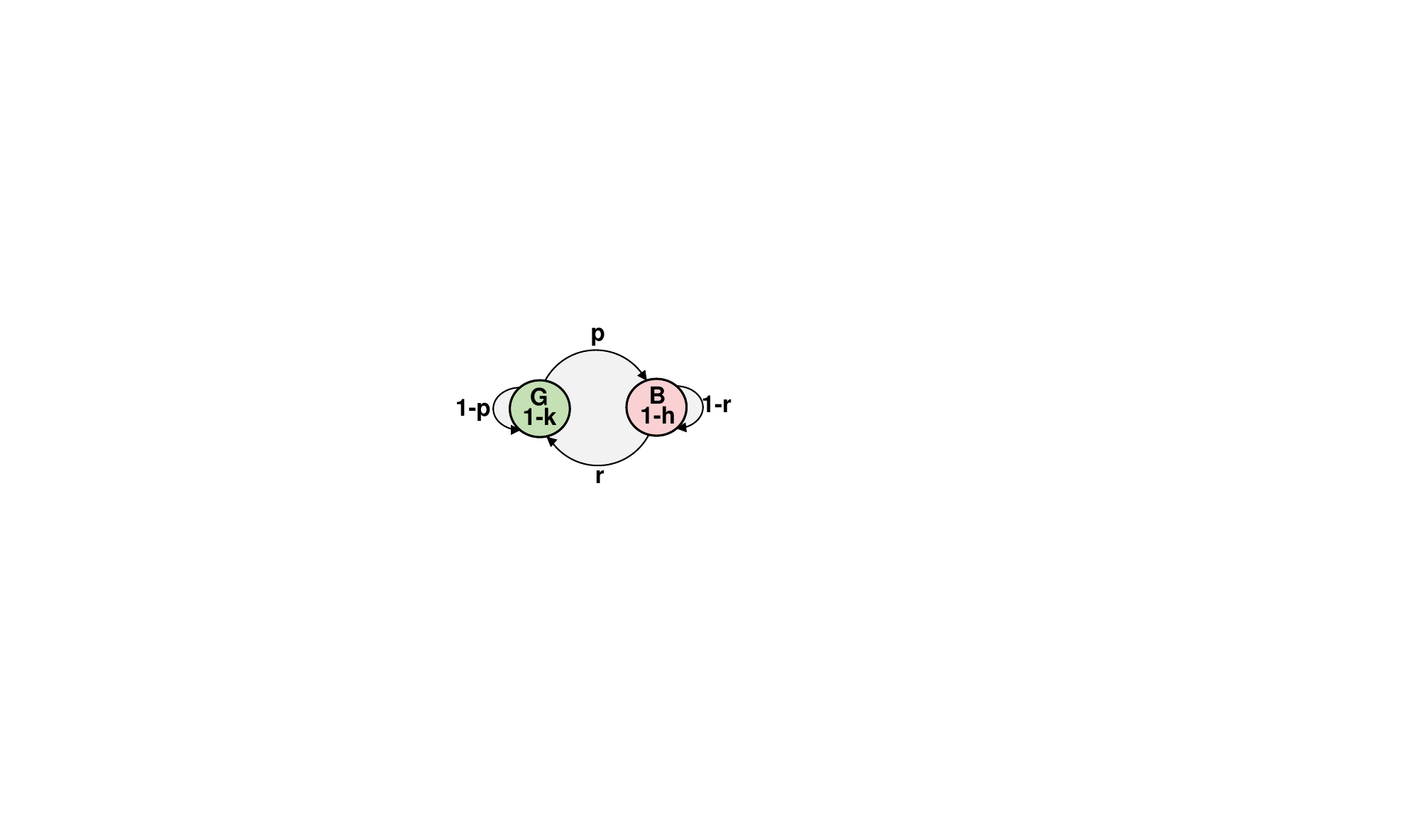}
\caption{{Gilbert-Elliot model.}}
\label{gemodelraw}
\end{figure}

\subsection{SCR's Impact on Entropy Coding}
We employ SCR to reduce the loss of entire channels due to packet loss in our channel-wise progressive coding. Rearranging the spatial and channel dimensions inevitably disrupts the original structure of the latent feature space, which affects the extraction of hyper-prior features using convolutional neural networks. However, in this scenario, features at adjacent positions become less correlated, which leads to a greater focus on the information at the corresponding positions during probability estimation. This aligns with the assumption of independence and identical distribution (i.i.d.) of such hyperprior model. Therefore, the impact of this process on compression performance is twofold, and experimental results in Fig.~\ref{fig:SCR_ablation} confirm that it does not significantly affect performance. 

\subsection{Experiments on ConvNeXt}
\begin{table*}[t]
\centering
\begin{tabular}{cc|ccc|ccc|cccc}
\toprule
\multirow{2}{*}{$p_e$} & \multirow{2}{*}{Method} & \multirow{2}{*}{bpp} & \multicolumn{2}{c|}{PSNR(dB)} & \multirow{2}{*}{bpp} & \multicolumn{2}{c|}{PSNR(dB)} & \multirow{2}{*}{bpp} & \multicolumn{2}{c}{PSNR(dB)} \\
\cline{4-5} \cline{7-8} \cline{10-11}
& & & $mean$ \(\uparrow\) & $var$ \(\downarrow\) & & $mean$ \(\uparrow\) & $var$ \(\downarrow\) & & $mean$ \(\uparrow\) & $var$ \(\downarrow\) \\
\midrule
&JPEG2000 & 0.137 & 24.201 & 0.133 & 0.205 & 24.517 & 0.269& 0.342 & 25.013 & 0.310  \\
\multirow{2}{*}{5\%}
&Baseline & 0.133 & 26.551 & 0.392 & 0.215 & 26.976 & 0.397 & 0.336 & 28.619 & 0.991 \\
&Random & 0.133 & 26.657 & 0.064 & 0.220 & 27.754 & 0.061 & 0.351 & 29.256 & 0.151  \\
&Ours & 0.131 & 26.850 & 0.015 & 0.210 & 28.247 & 0.043 & 0.338 & 29.561 & \textbf{0.007} \\
&Ours (ConvNeXt) & 0.123 & \textbf{27.302} & \textbf{0.008} & 0.201 & \textbf{28.843} & \textbf{0.037} & 0.310 & \textbf{30.088} & 0.018 \\
\midrule
&JPEG2000 & 0.137 & 23.448 & 0.220 & 0.205 & 23.626 & 0.226 & 0.342 & 23.966 & 0.318  \\
\multirow{2}{*}{10\%}&Baseline & 0.133 & 25.131 & 0.485 & 0.215 & 25.855 & 0.195 & 0.336 & 27.056 & 0.803 \\
&Random & 0.133 & 25.858 & 0.195 & 0.221 & 27.320 & 0.156 & 0.351 & 28.569 & 0.041  \\
&Ours & 0.131 & 26.342 & 0.061 & 0.210 & 27.605 & \textbf{0.100} & 0.342 & 28.801 & \textbf{0.007} \\
&Ours (ConvNeXt) & 0.123 & \textbf{26.747} & \textbf{0.054} & 0.201 & \textbf{28.062} & 0.149 & 0.310 & \textbf{29.509} & 0.061 \\
\midrule
&JPEG2000 & 0.137 & 22.530 & 0.304 & 0.205 & 22.524& 0.313& 0.342 & 22.564 & 0.337  \\
\multirow{2}{*}{20\%}&Baseline & 0.133 & 23.257  & 1.222  & 0.215 & 22.524 & 1.666 & 0.336 & 23.592 & 1.341\\
&Random &  0.133 & 25.217 & \textbf{0.071} & 0.228 &  25.886 & 0.192 & 0.358 & 27.620 & \textbf{0.016}    \\
&Ours & 0.131 & 25.578 & 0.108 & 0.217 & 26.617 & \textbf{0.043} & 0.346 & 27.966  & 0.043  \\
&Ours (ConvNeXt) & 0.122 & \textbf{26.177} & 0.113 & 0.195 & \textbf{27.314} & 0.106 & 0.305 & \textbf{28.329} & 0.128 \\
\bottomrule
\end{tabular}
\caption{Performance comparison of models under packet loss with a uniform distribution on the Kodak dataset, where $p_e$ represents the probability of packet loss, $mean$ and $var$ respectively represent the average PSNR and variance for 10 test samples. The bold values indicate the best results. The symbol $\downarrow$ signifies that a smaller value is better for that metric, while $\uparrow$ indicates that a larger value is preferable.}
\label{table:uniform_Convnext}
\end{table*}

\begin{table*}[htbp]
\centering
\begin{tabular}{cc|ccc|ccc|cccc}
\toprule
\multirow{2}{*}{$p_e$} & \multirow{2}{*}{Method} & \multirow{2}{*}{bpp} & \multicolumn{2}{c|}{PSNR(dB)} & \multirow{2}{*}{bpp} & \multicolumn{2}{c|}{PSNR(dB)} & \multirow{2}{*}{bpp} & \multicolumn{2}{c}{PSNR(dB)} \\
\cline{4-5} \cline{7-8} \cline{10-11}
& & & $mean$ \(\uparrow\) & $var$ \(\downarrow\) & & $mean$ \(\uparrow\) & $var$ \(\downarrow\) & & $mean$ \(\uparrow\) & $var$ \(\downarrow\) \\
\midrule
&JPEG2000 & 0.098 & 22.995 & 0.568 & 0.147 & 22.997 & 0.772 & 0.245 & 23.196 & 0.672  \\
\multirow{2}{*}{5\%}
&Baseline & 0.102 & 28.119 & 0.267 & 0.160 & 28.644 & 0.140 & 0.247 & 29.586 & 0.247 \\
&Random & 0.102 & 28.379 & 0.067 & 0.164 & 29.737 & 0.140 & 0.258 & 30.958 & 0.074  \\
&Ours & 0.100 & 28.795 & \textbf{0.039} & 0.158 & 30.074 & 0.060 & 0.252 & 31.154 & \textbf{0.062} \\
&Ours (ConvNeXt) & 0.095 & \textbf{29.030} & 0.096 & 0.155 & \textbf{30.592} & \textbf{0.046} & 0.231 & \textbf{31.525} & \textbf{0.062} \\
\midrule
&JPEG2000 & 0.098 & 20.747 & 0.435 & 0.147 & 20.667 & 0.474& 0.245 & 20.862 & 0.492  \\
\multirow{2}{*}{10\%}
&Baseline & 0.102 & 26.450 & 0.831 & 0.160 & 27.290 & 0.282 & 0.247 & 27.638 & 1.023 \\
&Random & 0.102 & 27.329 & 0.203 & 0.164 & 28.697 & 0.132 & 0.258 & 30.037 & 0.201  \\
&Ours & 0.100 & 28.261 & \textbf{0.115} & 0.158 & 29.210 & 0.200 & 0.252 & 30.491 & \textbf{0.080} \\
&Ours (ConvNeXt) & 0.095 & \textbf{28.567} & 0.186 & 0.155 & \textbf{29.925} & \textbf{0.089} & 0.231 & \textbf{31.047} & 0.103 \\
\midrule
&JPEG2000 & 0.098  & 18.385 & 0.418 & 0.147 & 18.934 & \textbf{0.260} & 0.245 & 18.729 & 0.576   \\
\multirow{2}{*}{20\%}
&Baseline & 0.102 & 23.460 & 1.776 & 0.160 & 23.994 & 1.991 & 0.247 & 24.714 & 0.840\\
&Random & 0.104 & 26.781 & 0.807 & 0.174 & 26.957 & 0.348 & 0.265 & 28.879 & 0.170  \\
&Ours & 0.103 & \textbf{27.316} & \textbf{0.110} & 0.167 & 27.873 & 0.318 & 0.257 & \textbf{29.469} & \textbf{0.122} \\
&Ours (ConvNeXt) & 0.095 & 27.003 & 0.310 & 0.155 & \textbf{28.105} & 0.343 & 0.231 & 29.404 & 0.336 \\
\bottomrule
\end{tabular}
\caption{Comparison of model performance under uniform packet loss on the CLIC professional validation dataset, where $p_e$ represents the probability of packet loss, $mean$ and $var$ respectively represent the average PSNR and variance for 10 test samples. The bold values indicate the best results. The symbol $\downarrow$ signifies that a smaller value is better for that metric, while $\uparrow$ indicates that a larger value is preferable.}
\label{table:uniform_CLIC}
\end{table*}

We also replaced the regular convolution with ConvNeXt to achieve more competitive performance. The experimental results are shown in the Table~\ref{table:uniform_Convnext}.

\subsection{Experiments on CLIC}
To further validate our approach, we conducted tests using the CLIC professional validation dataset, as shown in Table~\ref{table:uniform_CLIC}. Due to constraints on maximum packet size and the number of packets, GEO satellite networks are typically not ideal for transmitting high-resolution images from the CLIC dataset. Nonetheless, to demonstrate the effectiveness of our method across various image resolutions, we increased the maximum packet size to 4500 bytes for these experiments. The results indicate that our method consistently achieves the highest and most stable decoded image quality, even with high-resolution images.

\subsection{More Visualization Results}
\label{sec:appendix:extreme}
\begin{figure*}[htbp]
\centering
\includegraphics[width=1.8\columnwidth]{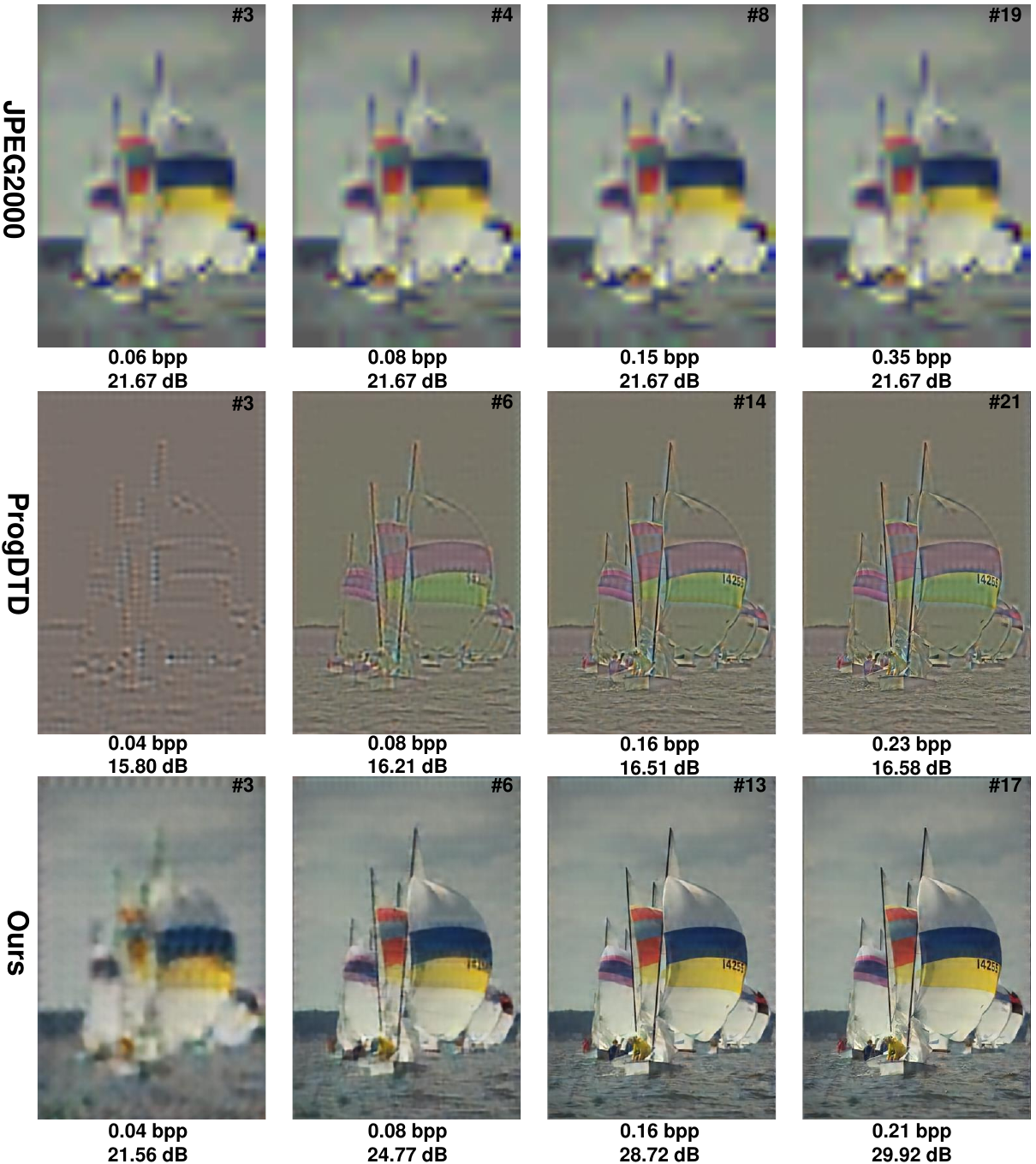}
\caption{Visualizations of progressive coding under packet loss on Kodim09. \# p denotes the reconstructed result after receiving the p-th packet at the decoder. Starting from the first column, all methods have lost the first two packets. Our method consistently demonstrates superior performance.}
\label{fig:extremesub}
\end{figure*}

\begin{figure*}[htbp]
\centering
\includegraphics[width=1.8\columnwidth]{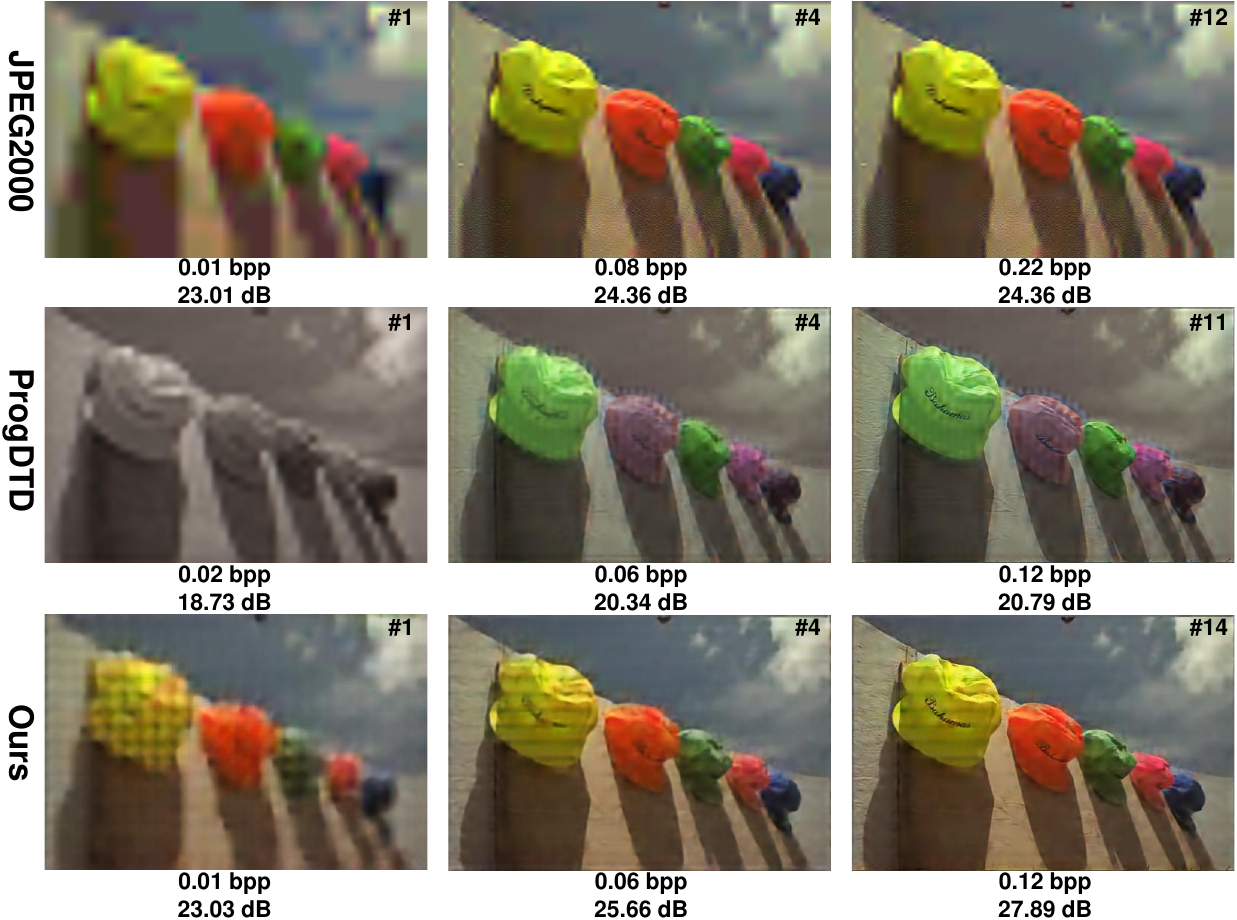}
\caption{Visualizations of progressive coding under packet loss on Kodim03. \# p denotes the reconstructed result after receiving the p-th packet at the decoder. The first column shows the initial results of progressive transmission without any packet loss for the three
methods. The results starting from the second column show the outcome after the 2nd packet is lost. Our method consistently demonstrates superior performance.}
\label{fig:sub2}
\end{figure*}

Fig.~\ref{fig:extremesub} and Fig.~\ref{fig:sub2} present additional visual results of progressive encoding under packet loss conditions.

\label{sec:appendix:packing}
\begin{algorithm}
\caption{Data Packing}
\begin{algorithmic}[1]
\REQUIRE $\hat{y}^r$: quantized latent feature
\REQUIRE $B_{max}$: maximum packet size
\ENSURE Each packet is smaller than $B_{max}$

\STATE $\text{Packets} \gets [ \  ]$
\STATE $\text{CurrentPacket} \gets [ \  ]$
\STATE $\text{CurrentSize} \gets 0$
\STATE $\text{EstSize} \gets \text{Estimate}(\hat{y})$ 

\FOR{$i = 1$ \TO $C$}
    \IF{$\text{CurrentSize} + \text{EstSize}_i \leq B_{max}$}
        \STATE \text{CurrentPacket.append(}$i$\text{)}
        \STATE $\text{CurrentSize} \gets \text{CurrentSize} + \text{EstSize}_i$
    \ELSE
        \STATE \text{Packets.append(}\text{CurrentPacket}\text{)}
        \STATE $\text{CurrentPacket} \gets [i]$
        \STATE $\text{CurrentSize} \gets \text{EstSize}_i$
    \ENDIF
\ENDFOR

\FOR{$i=1$ \textbf{to} \text{len(Packets)}}
    \STATE $\text{ActualSize} \gets \text{ComputeActualSize}(\text{Packets[$i$]})$ 
    \IF{$\text{ActualSize} > B_{max}$}
        \STATE $m$ = \text{len(Packets}[$i$]) $ // 2$
        \STATE $\text{Packets}[i] \to \text{Packets}[i][:m] \  \text{Packets}[i][m:] $
        \STATE $\text{Packets}[i]_{new} \gets \text{Packets}[i][:m]$
        \STATE $\text{Packets.insert}(i + 1,  \text{Packets}[i][m:])$
    \ENDIF
\ENDFOR

\FOR{$i = 1$ \TO \text{len(}\text{Packets}\text{)}}
    \IF{\text{Packets[$i$][$-1$].index} \text{ mod } 4 == 0}
        \STATE \text{Packets[$i+1$].insert(1, packet.pop())}
    \ENDIF
\ENDFOR

\RETURN Packets

\end{algorithmic}
\label{al:packet}
\end{algorithm}

\end{document}